\documentclass[runningheads]{llncs}
\usepackage[T1]{fontenc}
\usepackage{graphicx}

\usepackage{amssymb}
\usepackage{amsmath}
\usepackage{multirow}
\usepackage{makecell}
\usepackage{xcolor}
\usepackage{subcaption}

\begin{document}
	\title{Memory-Efficient 3D High-Resolution Medical Image Synthesis Using CRF-Guided GANs}
	\titlerunning{CRF-Guided GANs For Memory-Efficient Image Synthesis}
	\author{Mahshid Shiri\inst{1}\orcidID{0009-0008-5780-724X} \and
		Alessandro Bruno\inst{2}\orcidID{0000-0003-0707-6131} \and
		Daniele Loiacono\inst{1}\orcidID{0000-0002-5355-0634}}
	\authorrunning{M. Shiri et al.}
	\institute{Politecnico di Milano, Department of Electronics, Information, and Bioengineering, Piazza Leonardo da Vinci, 32, 20133 Milan, Italy. \\ \and
		IULM University, Department of Business, Law, Economics, and Consumer Behaviour 'Carlo A. Ricciardi'. Via Carlo Bo, 1. 20143, Milan, Italy.}
	\maketitle              % typeset the header of the contribution
	\begin{abstract}
		Generative Adversarial Networks (GANs) have many potential medical imaging applications. Due to the limited memory of Graphical Processing Units (GPUs), most current 3D GAN models are trained on low-resolution medical images, these models cannot scale to high-resolution or are susceptible to patchy artifacts. 
		In this work, we propose an end-to-end novel GAN architecture that uses Conditional Random field(CRF) to model dependencies so that it can generate consistent 3D medical Images without exploiting memory. To achieve this purpose, the generator is divided into two parts during training, the first part produces an intermediate representation and CRF is applied to this intermediate representation to capture correlations. The second part of the generator produces a random sub-volume of image using a subset of the intermediate representation. This structure has two advantages: first, the correlations are modeled by using the features that the generator is trying to optimize. Second, the generator can generate full high-resolution images during inference. Experiments on Lung CTs and Brain MRIs show that our architecture outperforms state-of-the-art while it has lower memory usage and less complexity.
		
	\end{abstract}
	\section{Introduction}
		Generative models have gained considerable popularity among researchers for a variety of tasks, including data augmentation\cite{celard2023survey}. Generative Adversarial Networks (GANs)\cite{goodfellow2014generative} have made remarkable strides in the biomedical field. These advancements are largely due to recent improvements in computational power and the availability of extensive medical datasets.
		
		\indent 3D GAN models were proposed for reducing noise in low-dose CT scans\cite{shan20183}, enhancing the quality of CT images\cite{kudo2019virtual}, generating realistic 3D brain MRI images\cite{jin2019applying}, and creating tumor masks for segmentation\cite{cirillo2021vox2vox}. However, these methods have a limitation in the generated image resolution, which is usually $128^3$ or smaller, because of limited memory during training\cite{sun2022hierarchical}.
		
		\indent Memory-efficient GANs aim to balance memory efficiency and generative capability, ensuring that GANs can be trained effectively even with limited computational resources. To this aim, the experimental campaigns by Lei et al.\cite{lei2019mri} and Yu et al.\cite{yu20183d} focused on producing new images slice by slice or patch by patch. However, as these methods generate patches and slices independently, the related boundaries may exhibit artefacts. Also, Uzunova et al. [23] use two GANs to generate images, with the first network releasing a lower-resolution version of the image and the second producing higher-resolution patches conditioned on the first GAN. This approach is still patch-based and lacks a comprehensive understanding of the entire image structure. Furthermore, it does not support end-to-end training, making it impractical to use encoders in this context.
		
		\indent Also, Sun et al.\cite{sun2022hierarchical} proposed an end-to-end hierachichal GAN architecture (HA-GAN), capable of generating high-resolution 3D images at a resolution of $256^3$. The HA-GAN architecture consists of two interconnected GANs: low-resolution GAN that produces a low-resolution version of the 3D image, capturing essential global structure with reduced computational and memory requirements, and high-resolution GAN that generates high-resolution patches for a randomly selected sub-volume of the image. The high-resolution generator shares its initial layers with the low-resolution generator, ensuring consistency and efficient use of memory. During inference, HA-GAN is capable of directly generating full high-resolution 3D images without the need for additional post-processing. HA-GAN demonstrates superior performance compared to state-of-the-art models in generating high-resolution 3D images.
		
		\indent This work aims to address the limitations of existing 3D GAN models by presenting a novel architecture for high-resolution 3D medical image synthesis that combines Conditional Random Fields (CRF) with GANs. The main contribution of this research is the development of a more memory-efficient and high-performing architecture. This study aims to advance the quality and applicability of synthetic medical images by tackling key challenges such as consistency, memory efficiency, model complexity, and performance.
		
	\section{Conditional Random Field(CRF)}
			Conditional Random Field(CRF) is a type of probabilistic graphical model used in machine learning. It tackles dependencies between the output variables, taking into account the data's sequential or structural nature. CRFs˝ allow capturing correlations between image patches. Let's denote the embedding of patches as \( x = \{x_i\}_{i=1}^N \), where \( N \) represents the number of patches. Each patch \( i \) is associated with a random variable \( y = \{y_i\}_{i=1}^N \), denoting the binary label of each patch. The conditional distribution \( P(Y | x) \) can be formulated as a CRF with a Gibbs distribution, as depicted in Equation \ref{crfgib}.
			\begin{equation}
				P(Y = y \mid x) = \frac{1}{Z(x)} \exp\left( -E(y, x) \right)
			\label{crfgib}
			\end{equation}
			\indent Here, \( E(y, x) \) represents the energy function, quantifying the cost of \( Y \) assuming a specific configuration \( y \) given \( x \). \( Z(x) \) stands for the partition function ensuring that \( P(Y = y | x) \) is a legitimate probability distribution. In the context of a fully-connected pairwise CRF\cite{krahenbuhl2011efficient}, the energy function can be expressed as Equation \ref{energy}.
			\begin{equation}
				E(y, x) = \sum_{i} u(y_i) + \sum_{i<j} p(y_i, y_j)
			\label{energy}
			\end{equation}
			\indent \( \sum_{i} \) and \( \sum_{i<j} \) indicate summations over all patches and pairwise combinations of patches, respectively. \( u(y_i) \) signifies the unary potential associated with patch \( i \)  that measures the cost of patch $i$ taking the label $y_i$ given the patch embedding $x_i$, while \( p(y_i, y_j) \) represents the pairwise potential between patches \( i \) and \( j \) that  measures the cost of jointly assigning patch $i, j$ with labels $y_i,y_j$ given the patch embeddings $x_i,x_j$ respectively. Pairwise potential $p (y_i,y_j)$ models spatial correlations between neighboring patches, and would encourage low cost for assigning $y_i,y_j$ with the same label if $x_i,x_j$ are similar\cite{wallach2004conditional}.
		
			\indent Li et al.\cite{li2018cancer} incorporate CRF on top of a feature extractor to refine segmentation results and ensure spatial coherence within the image. Zhong\cite{zhong2019spectral} explores the combined use of GANs and CRF for hyper-spectral image classification.  Zhong\cite{zhong2019spectral} employed CRFs as a post processing layer to refine the classification results obtained from GAN-generated hyper-spectral images.
			
	\section{Method}
		In this section, we explain the underlying idea behind the proposed method and the overall architecture.
		\subsection{Idea Behind Proposed Architecture}
			When training a GAN, the generator learns different levels of detail across its layers. The initial layers focus on broad, coarse features, while the later layers refine the finer details. The generator starts with a noise vector, progressively adding information with each layer, until the final output is a complete image.
			
			\indent We can split the generator into two parts: the first part generates an embedding that captures the overall structure of the image, and the second part refines this into the full image. The point in the network where we divide the network is critical because the earlier layers are too abstract, and the later layers are too specific. A good division is usually somewhere in the middle, where the output is detailed enough to be informative.
			
			\indent From another perspective, the output of the first layers can be seen as the embedding of the images, which encapsulates the knowledge of the global structure. A common approach in the literature\cite{sun2022hierarchical,uzunova2019multi} to provide more feedback on the first layers regarding the consistency in the final images is to use another GAN. In this work, we capture structures and dependencies by applying CRFs to the embeddings rather than training an additional GAN.
			
			\indent CRFs are lightweight discriminative models that capture correlations between patches without producing extra outputs. They can be integrated into a simple GAN architecture to provide feedback on the consistency and structures between patches to the generator by analyzing embeddings and their neighbors. Meanwhile, the discriminator assesses how realistic the generated images are by examining patches of the image. This dual feedback mechanism ensures that the generator receives feedback on both consistency and realism.
		\subsection{Components}
			\subsubsection{Two-Step GAN Structure:}
				As depicted in Fig~.\ref{crfgan}, the generator is structured into two sections to access image embeddings. The initial section $(G_1)$ generates an embedding of images $(A)$ from a noise variable $(z)$. During training, a random subset of the embedding $A$ is selected $(A_r)$ and passed to the second network. During inference, the whole embedding $A$ is passed to the second network to generate full image. The second network $(G_2)$ generates patches of image in full resolution from the selected subset $A_r$. This selection procedure is also applied to real images ($X$) prior to their inclusion in the discriminator input. Consequently, the loss function of Generators and Discriminator (D) would be as Equation \ref{eqTwoStepG}.
				\noindent
				\begin{equation}\label{eqTwoStepG}
					\centering
					\begin{aligned}
						L_{GAN}(G_1,G_2,D) = &\min_{G_1,G_2} \max_{D}[ \mathbb{E}_{r \sim p_r}[ \mathbb{E}_{x \sim p_{data}}[\log D(X_{(r:r+constant)})] \\
						&+ \mathbb{E}_{z \sim p_z}[ \log\big(1 - D(G_2(G_1(z))_{(r:r+constant)}))]]]
					\end{aligned}
				\end{equation}
			\subsubsection{Half-Encoder Structure:}
				An encoder is integrated into the architecture to acquire a condensed representation of the entire image and prevent mode collapse and it takes patches of images as input. Despite normal encoders, the designed encoder only reverses the procedure of $G_2$, so it is referred to as the half-encoder. The output of the encoder is then fed into $G_2$ for training and subsequently compared with the original patches. The designed encoder is able to generate embedding of both synthetic and real images which is going to be used during training of CRF component. The loss is calculated as Equation \ref{eq_half_enc}, where $l_1$ loss is utilized to achieve sharper results compared to $l_2$ loss\cite{zhu2017unpaired}.
				\noindent
				\begin{equation}\label{eq_half_enc}
					L_{\text{reconstruct}}(E) = \min_{E} [ \mathbb{E}_{(x \sim p_{\text{data}}, r \sim p_r)} \| X_r - G_2(\widehat{A_r})\|_1 ] \text{,where} \widehat{A_r} = E(X_r)
				\end{equation}
		\subsection{Proposed Architecture (CRF-GAN)}
			\begin{figure}
				\includegraphics[width=\textwidth]{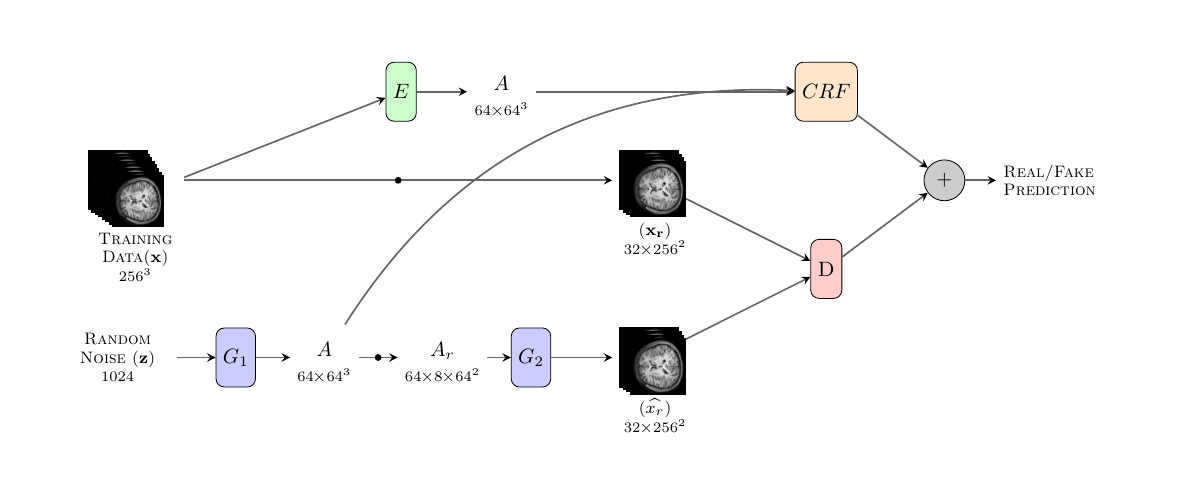}
				\caption{CRF-GAN architecture. CRF is applied to the embedding of real images(output of $E$) and synthetic images(output of $G_1$)}
				\label{crfgan}
			\end{figure}
			By utilizing the two-step GAN and half-encoder, the embedding of real and synthetic images is provided. As said earlier, this embedding($A$) does not illustrate details but contains the general structure of images. To use the information that is encoded into $A$, CRF is applied to the output of $G_1$($A$) and captures correlations. So, the first section of the generator would receive feedback both from the CRF and the discriminator. In this way, the signal about the general structure is transferred to the generator so that the inconsistency between patches is prevented.
			
			\indent This approach has several advantages. First, the CRF is trained using the same features produced by the generator, secondly, training a CRF incurs significantly less overhead compared to training an extra GAN, resulting in a model with lower complexity than existing memory-efficient GAN architectures. Equation \ref{eq_crfgan} defines the loss function, and Fig.~\ref{crfgan} illustrates the architecture.
			\noindent\begin{equation}\label{eq_crfgan}
				\begin{split}
					&\mathcal{L}_{\text{CRF-GAN}}(G_1,G_2,D,CRF) = \\
					&\min_{G_1,G_2} \max_{D,CRF} [ \mathbb{E}_{r \sim p_r}[ \mathbb{E}_{x \sim p_{data}}[ \frac{\log(D(X_{r:r+constant}) + CRF(E(X)))}{2}] \\
					&\quad + \mathbb{E}_{z \sim p_z}[ \frac{\log(1 - (D(G_2(G_1(z))_{r:r+\text{constant}}) + CRF(G_1(z))))}{2}]]]
				\end{split}
			\end{equation}
	\section{Experiments}
			HA-GAN\cite{sun2022hierarchical} has demonstrated superior performance compared to baseline methods such as WGAN\cite{gulrajani2017improved}, VAE-GAN\cite{larsen2016autoencoding}, $\alpha$GAN\cite{kwon2019generation}, ProgressiveGAN\cite{karras2017progressive}, 3D StyleGAN 2\cite{hong20213d}, and CCE-GAN\cite{xing2021cycle}. Additionally, HA-GAN\cite{sun2022hierarchical} is the only model capable of generating images with a size of $256^3$ due to memory constraints. Therefore, the proposed architecture is compared to HA-GAN in this section.
		
\begin{figure}
				\centering
				\begin{minipage}[t]{\linewidth}
					\centering
					\begin{subfigure}[b]{0.14\textwidth}
						\centering
						\includegraphics[width=\textwidth]{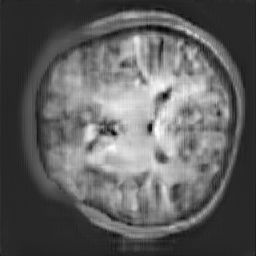}
						\caption*{20000}
					\end{subfigure}\hfill
					\begin{subfigure}[b]{0.14\textwidth}
						\centering
						\includegraphics[width=\textwidth]{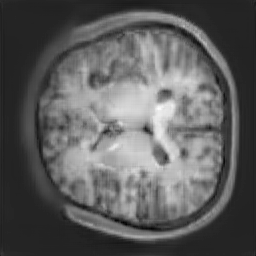}
						\caption*{30000}
					\end{subfigure}\hfill
					\begin{subfigure}[b]{0.14\textwidth}
						\centering
						\includegraphics[width=\textwidth]{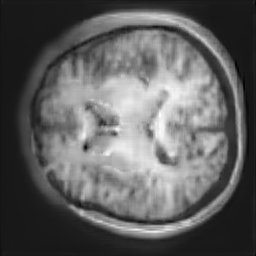}
						\caption*{40000}
					\end{subfigure}\hfill
					\begin{subfigure}[b]{0.14\textwidth}
						\centering
						\includegraphics[width=\textwidth]{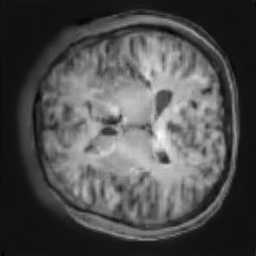}
						\caption*{50000}
					\end{subfigure}\hfill
					\begin{subfigure}[b]{0.14\textwidth}
						\centering
						\includegraphics[width=\textwidth]{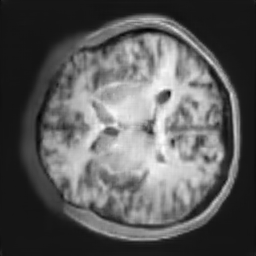}
						\caption*{60000}
					\end{subfigure}\hfill
					\begin{subfigure}[b]{0.14\textwidth}
						\centering
						\includegraphics[width=\textwidth]{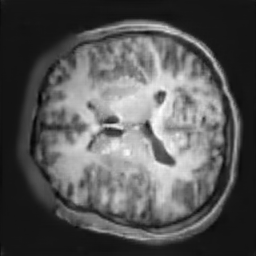}
						\caption*{70000}
					\end{subfigure}\hfill
					\begin{subfigure}[b]{0.14\textwidth}
						\centering
						\includegraphics[width=\textwidth]{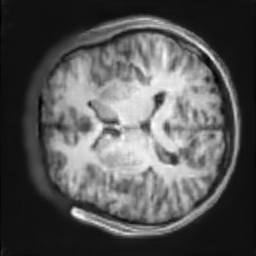}
						\caption*{80000}
					\end{subfigure}
					\caption*{CRF-GAN synthetic images at different iterations}
				\end{minipage}
				
				\begin{minipage}[t]{\linewidth}
					\centering
					\begin{subfigure}[b]{0.14\textwidth}
						\centering
						\includegraphics[width=\textwidth]{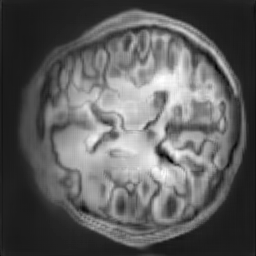}
						\caption*{20000}
					\end{subfigure}\hfill
					\begin{subfigure}[b]{0.14\textwidth}
						\centering
						\includegraphics[width=\textwidth]{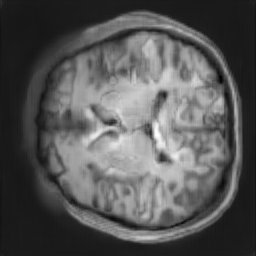}
						\caption*{30000}
					\end{subfigure}\hfill
					\begin{subfigure}[b]{0.14\textwidth}
						\centering
						\includegraphics[width=\textwidth]{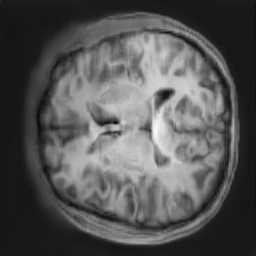}
						\caption*{40000}
					\end{subfigure}\hfill
					\begin{subfigure}[b]{0.14\textwidth}
						\centering
						\includegraphics[width=\textwidth]{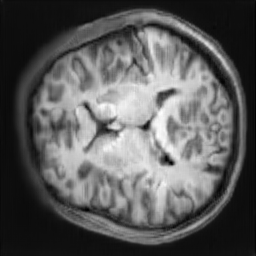}
						\caption*{50000}
					\end{subfigure}\hfill
					\begin{subfigure}[b]{0.14\textwidth}
						\centering
						\includegraphics[width=\textwidth]{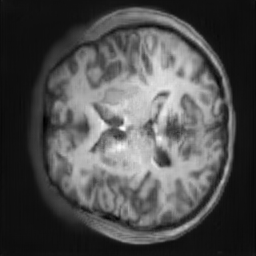}
						\caption*{60000}
					\end{subfigure}\hfill
					\begin{subfigure}[b]{0.14\textwidth}
						\centering
						\includegraphics[width=\textwidth]{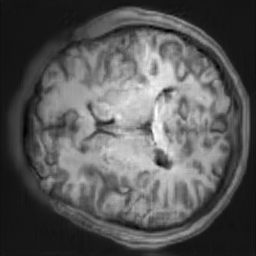}
						\caption*{70000}
					\end{subfigure}\hfill
					\begin{subfigure}[b]{0.14\textwidth}
						\centering
						\includegraphics[width=\textwidth]{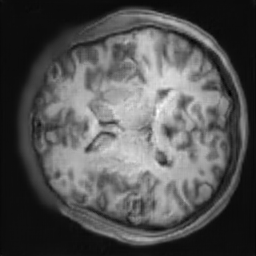}
						\caption*{80000}
					\end{subfigure}
					\caption*{HA-GAN synthetic images at different iterations}
				\end{minipage}
				\caption{Synthetic images of CRF-GAN and HA-GAN at different iterations of training on GSP dataset}
				\label{fig_series_GSP_photos}
			\end{figure}

\begin{figure}
				\centering
				\begin{minipage}[t]{\linewidth}
					\centering
					\begin{subfigure}[b]{0.14\textwidth}
						\centering
						\includegraphics[width=\textwidth]{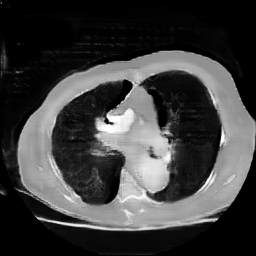}
						\caption*{20000}
					\end{subfigure}\hfill
					\begin{subfigure}[b]{0.14\textwidth}
						\centering
						\includegraphics[width=\textwidth]{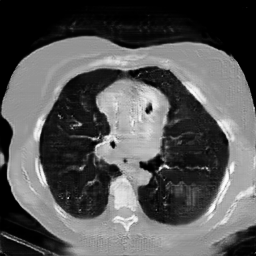}
						\caption*{30000}
					\end{subfigure}\hfill
					\begin{subfigure}[b]{0.14\textwidth}
						\centering
						\includegraphics[width=\textwidth]{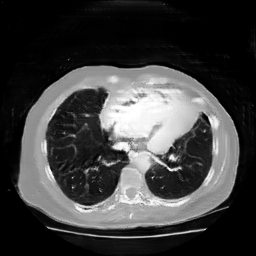}
						\caption*{40000}
					\end{subfigure}\hfill
					\begin{subfigure}[b]{0.14\textwidth}
						\centering
						\includegraphics[width=\textwidth]{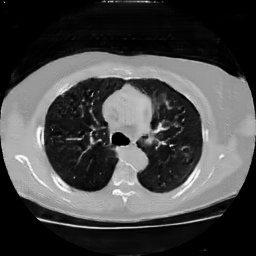}
						\caption*{50000}
					\end{subfigure}\hfill
					\begin{subfigure}[b]{0.14\textwidth}
						\centering
						\includegraphics[width=\textwidth]{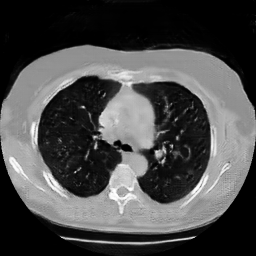}
						\caption*{60000}
					\end{subfigure}\hfill
					\begin{subfigure}[b]{0.14\textwidth}
						\centering
						\includegraphics[width=\textwidth]{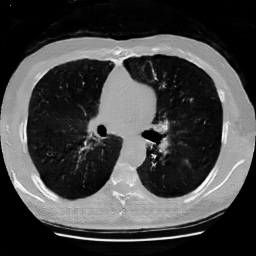}
						\caption*{70000}
					\end{subfigure}\hfill
					\begin{subfigure}[b]{0.14\textwidth}
						\centering
						\includegraphics[width=\textwidth]{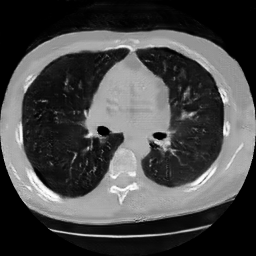}
						\caption*{80000}
					\end{subfigure}
					\caption*{CRF-GAN synthetic images at different iterations}
				\end{minipage}
				
				\begin{minipage}[t]{\linewidth}
					\centering
					\begin{subfigure}[b]{0.14\textwidth}
						\centering
						\includegraphics[width=\textwidth]{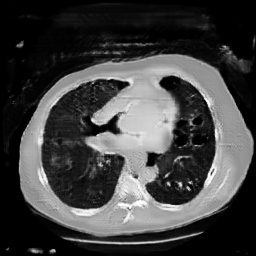}
						\caption*{20000}
					\end{subfigure}\hfill
					\begin{subfigure}[b]{0.14\textwidth}
						\centering
						\includegraphics[width=\textwidth]{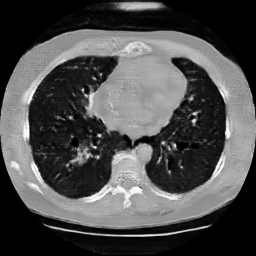}
						\caption*{30000}
					\end{subfigure}\hfill
					\begin{subfigure}[b]{0.14\textwidth}
						\centering
						\includegraphics[width=\textwidth]{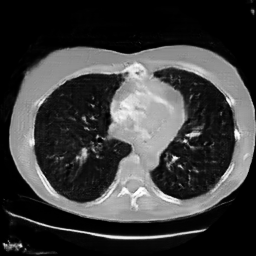}
						\caption*{40000}
					\end{subfigure}\hfill
					\begin{subfigure}[b]{0.14\textwidth}
						\centering
						\includegraphics[width=\textwidth]{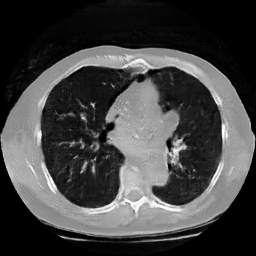}
						\caption*{50000}
					\end{subfigure}\hfill
					\begin{subfigure}[b]{0.14\textwidth}
						\centering
						\includegraphics[width=\textwidth]{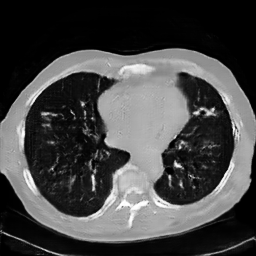}
						\caption*{60000}
					\end{subfigure}\hfill
					\begin{subfigure}[b]{0.14\textwidth}
						\centering
						\includegraphics[width=\textwidth]{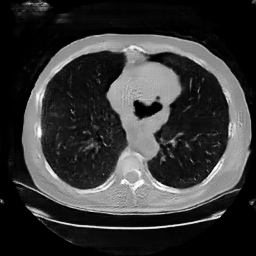}
						\caption*{70000}
					\end{subfigure}\hfill
					\begin{subfigure}[b]{0.14\textwidth}
						\centering
						\includegraphics[width=\textwidth]{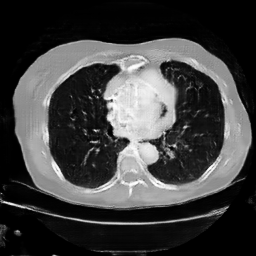}
						\caption*{80000}
					\end{subfigure}
					\caption*{HA-GAN synthetic images at different iterations}
				\end{minipage}
				\caption{Synthetic images of CRF-GAN and HA-GAN at different iterations of training on the LUNA16 dataset}
				\label{fig_series_LUNA_photos}
			\end{figure}

			\indent The study utilizes three 3D datasets, namely the GSP dataset\cite{DVN/25833_2014}, the LIDC-IDRI dataset\cite{armato2011lung}, and the dataset employed for the Lung Nodule Analysis 2016 (LUNA16) challenge\cite{setio2017validation} which is a subset of LIDC-IDRI dataset. Samples from generated images at different stages of training are presented in Fig.~\ref{fig_series_GSP_photos} and Fig.\ref{fig_series_LUNA_photos}.
				
			\subsection{Performance Analysis}
			Within this segment, we outline the selected assessment criteria employed to evaluate the efficiency of the suggested architecture.
				\subsubsection{FID and MMD Scores:}
				For the purpose of assessment, the metrics of Fréchet Inception Distance (FID)\cite{heusel2017gans} and Maximum Mean Discrepancy (MMD)\cite{gretton2012kernel} are computed. FID focuses on quantifying the dissimilarity between the generated and real distributions and captures both precision and recall. On the other hand, MMD differentiate between generated and real images with minimal sample and computational complexity\cite{borji2019pros}. Lower FID and MMD scores show the proximity of generated images to real images. 
				
				\indent To evaluate the performance, a subset comprising $20\%$ of the GSP dataset and $10\%$ of the LUNA16 dataset was designated as test data and the experiments are done on $256^3$ resolution. The resulting scores are tabulated in Table \ref{table_fid_mmd}. Notably, CRF-GAN achieved better FID and superior MMD scores across both datasets, underscoring its efficacy in generating images closely resembling real images.
				\begin{table}[htbp]
					\caption{FID and MMD scores computed on the $256^3$ resolution images of GSP and LUNA16 dataset. As lower FID and MMD scores outline better image fidelity, it can be seen that CRF-GAN achieved better results.}
					\label{table_fid_mmd}
					\centering
					\begin{tabular}{|l|l|l|l|l|}
						\hline
						\multirow{2}{*}{\textbf{Models}} & \multicolumn{2}{c|}{\textbf{GSP dataset}} & \multicolumn{2}{c|}{\textbf{LUNA16 dataset}} \\
						\cline{2-5}
						& \textbf{FID $\downarrow$} & \textbf{MMD $\downarrow$} & \textbf{FID $\downarrow$} & \textbf{MMD $\downarrow$} \\
						\hline
						HA-GAN & $0.009278$ & $0.003802$ & $0.061021$ & $0.086461$ \\ 
						\hline
						CRF-GAN & \boldmath{$0.007044$} & \boldmath{$0.000099$} & \boldmath{$0.047062$} & \boldmath{$0.084015$} \\ 
						\hline
					\end{tabular}
				\end{table}
				\subsubsection{Simplicity:}
					We explored the simplicity of CRF-GAN and HA-GAN models by analyzing the total number of learnable parameters at different resolutions, which are presented in Table \ref{tab_param}. The proposed architecture demonstrates a reduction in parameters at each resolution, with decreases of 27.4\%, 28.1\%, and 28.2\% observed for resolutions \(64^3\), \(128^3\), and \(256^3\) respectively. 
				\begin{table}[htbp]
					\caption{The total number of learnable parameters for CRF-GAN and HA-GAN at varying resolutions in millions (M).}
					\label{tab_param}
					\centering
					\begin{tabular}{|l|l|l|}
						\hline
						\multirow{2}{*}{\textbf{Resolution}} & \multicolumn{2}{c|}{\textbf{\#Parameters}} \\
						\cline{2-3}
						& \textbf{CRF-GAN} & \textbf{HA-GAN} \\ 
						\hline
						$64^3$ & $57.07$ M & $78.7$ M \\ 
						\hline
						$128^3$ & $57.20$ M & $79.58$ M \\ 
						\hline
						$256^3$ & $57.24$ M & $79.74$ M \\ 
						\hline
					\end{tabular}
				\end{table}
				\subsubsection{Memory Efficiency:}
					We captured the maximum memory allocated during training. This procedure was performed on batch sizes of 2, 4 and 6 on both $128^3$ and $256^3$ resolutions. The results are outlined in Table \ref{tab_memory_usage}. HA-GAN consistently used more memory than CRF-GAN at both $128^3$ and $256^3$ resolutions across all batch sizes. The memory usage disparity increased with higher batch sizes, with HA-GAN averaging about 9.4\% more memory usage than CRF-GAN.
				\begin{table}[htbp]
					\caption{Memory usage of HA-GAN and CRF-GAN for different batch sizes at $128^3$ and $256^3$ resolutions."--" indicates that the models exploit GPU's memory.}
					\label{tab_memory_usage}
					\centering
					\begin{tabular}{|l|l|l|l|l|l|l|}
						\hline
						\multirow{3}{*}{Models} & \multicolumn{6}{c|}{Max Memory allocated (MB)} \\
						\cline{2-7}
						& \multicolumn{3}{c|}{$128^3$} & \multicolumn{3}{c|}{$256^3$} \\
						\cline{2-7}
						& 2 & 4 & 6 & 2 & 4 & 6 \\
						\hline
						HA-GAN & 2812 & 2984 & 3098 & 7710 & 10434 & -- \\
						\hline
						CRF-GAN & 2622 & 2712 & 2774 & 7494 & 9010 & -- \\
						\hline
					\end{tabular}
				\end{table}
					
				\subsubsection{Training Speed:}
					In addition, we measured the number of iterations per second during training, setting the batch size to 2. CRF-GAN is approximately 14.6\%, 10.2\% faster than HA-GAN at $128^3$, $256^3$ spatial resolutions respectively, as shown in Table \ref{tab_speed}.
					\begin{table}[htbp]
						\caption{Training speed (iter/sec) for CRF-GAN and HA-GAN.}
						\label{tab_speed}
						\centering
						\begin{tabular}{|l|l|l|}
							\hline
							\multirow{2}{*}{\textbf{Resolution}} & \multicolumn{2}{c|}{\textbf{Training speed (iter/sec)}} \\
							\cline{2-3}
							& \textbf{CRF-GAN} & \textbf{HA-GAN} \\ 
							\hline
							$128^3$ & $1.394$ & $1.216$  \\ 
							\hline
							$256^3$ & $4.236$ & $3.845$  \\ 
							\hline
						\end{tabular}
					\end{table}	
				\subsubsection{Augmentation Task:}
				Frid-Adar et al.\cite{frid2018gan} demonstrated that GAN-generated samples enhance training dataset diversity and improve classifier's performance. To investigate the potential of synthetic images, we extended our CRF-GAN architecture for conditional image generation and designed a binary classification task. To this end, we used a subset of the LIDC-IDRI dataset, categorizing images into two classes: positive (nodules rated 5, high malignancy likelihood) and negative (no nodules). 	
			
				\indent The classifier was trained under three configurations: 1) using only real data, 2) using a 50-50 mix of real and CRF-GAN synthetic images, and 3) using a 50-50 mix of real and HA-GAN synthetic images. Each configuration was trained with 10 different seeds, and the average performance metrics are shown in Table \ref{tab_augm}. The dataset was imbalanced, with a higher number of scans containing nodules compared to those without.
			
				\indent To compare both architectures more fairly, we allowed the classifier to learn until early stopping was triggered based on the validation loss, rather than training the classifiers for a fixed number of iterations as done in\cite{sun2022hierarchical}. This approach lets the classifier use the provided data until performance plateaus, allowing us to better assess the impact of adding synthetic images.
		
				\indent The classifier with CRF-GAN synthetic images achieved the highest recall (79.99\%), indicating lower false negative cases. Precision values were high for both CRF-GAN (78.55\%) and HA-GAN (79.02\%), showing low false positive rates. The CRF-GAN classifier had the highest F1 score (79.21\%), indicating a good balance between precision and recall. The standard deviations for the precision, recall, and F1 scores across all configurations indicate variability in the model's performance, with CRF-GAN showing lower variability compared to HA-GAN, suggesting more consistent results.
			
				\indent In medical imaging, especially with critical conditions like nodules, it is crucial to prioritize models with higher recall to ensure fewer missed diagnoses. CRF-GAN's strength lies in its higher recall and F1 score, indicating that the synthetic images generated by CRF-GAN enhance the model's ability to detect nodules, despite some increase in false positives. This is generally preferable in medical imaging, where false negatives are more critical than false positives. On the other hand, HA-GAN's high precision but low recall and F1 score suggest that while the synthetic images from HA-GAN help avoid false positives, they fail to improve the model's ability to detect nodules.	
			\begin{table}[htbp]
				\caption{Average classifier performance trained with real data and, real $+$ synthetic data of CRF-GAN and real $+$ synthetic data of HA-GAN.}
				\label{tab_augm}
				\centering
				\begin{tabular}{|c|c|c|c|}
					\hline
					\textbf{Configuration} & \textbf{Precision (\%)} & \textbf{Recall (\%)} & \textbf{F1 Score (\%)} \\
					\hline
					Baseline & 76.97 $\pm$\scriptsize{2.34} & 77.27 $\pm$\scriptsize{9.06} & 76.92 $\pm$\scriptsize{5.27} \\
					\hline
					With CRF-GAN & 78.55 $\pm$\scriptsize{1.99}  & 79.99 $\pm$\scriptsize{5.19} & 79.21 $\pm$\scriptsize{3.19} \\
					\hline
					With HA-GAN & 79.02 $\pm$\scriptsize{2.42} & 73.33 $\pm$\scriptsize{7.8} & 75.79 $\pm$\scriptsize{3.63} \\
					\hline
				\end{tabular}
			\end{table}		
	\section{Conclusion}
	We presented a novel memory-efficient GAN architecture that showcases significant progress in generating high-resolution 3D medical images. By utilizing CRF to model dependencies within the intermediate representations generated by the generator, this architecture effectively addresses the limitations imposed by the limited memory capacity of GPUs. 

	\indent The experimental findings on Lung CTs and Brain MRIs confirm the advantage of this approach compared to state-of-the-art methods, highlighting its lower memory usage, reduced complexity, and improved performance of existing models. However, the incorporation of CRF contributes to smoother results that require further attention in future studies. 

	To evaluate the proposed model, we utilized both FID and MMD scores. FID is a well-known metric for measuring the quality of generated images, but its reliability can be limited in certain situations \cite{jayasumana2024rethinking}. For this reason, we also incorporated the MMD score as an additional evaluation metric. Furthermore, we examined the effectiveness of our model through data augmentation, which showed enhanced performance compared to existing models.
		
	We also conducted a preliminary evaluation with medical experts to assess the clinical realism of the generated images. The initial feedback was encouraging, and we plan to expand on this analysis in future work.

	\bibliographystyle{splncs04}
	\bibliography{Thesis_bibliography}

\end{document}